\documentclass{article}

\usepackage[preprint]{neurips_2020}

\usepackage{amsmath,amsfonts,bm}

\def\eqref#1{equation~\ref{#1}}

\def\1{\bm{1}}

\DeclareMathAlphabet{\mathsfit}{\encodingdefault}{\sfdefault}{m}{sl}
\SetMathAlphabet{\mathsfit}{bold}{\encodingdefault}{\sfdefault}{bx}{n}

\usepackage{neurips_2020}
\usepackage{graphicx}
\usepackage{subfigure}
\usepackage[utf8]{inputenc} 
\usepackage[T1]{fontenc}    
\usepackage{hyperref}       
\usepackage{url}            
\usepackage{booktabs}       
\usepackage{amsfonts}       
\usepackage{nicefrac}    
\usepackage{amsmath}
\usepackage{microtype}      
\newcommand{\mask}{\bm{S}}
\newcommand{\M}{\bm{M}}
\newcommand{\Mhat}{\bm{X}}

\newcommand{\bb}[1]{\bm{#1}}
\newcommand{\Rphi}{\bb{P}}
\newcommand{\Rpsi}{\bb{Q}}
\newcommand{\Phibold}{\mathbf{\Phi}}
\newcommand{\Psibold}{\mathbf{\Psi}}

\newcommand{\trace}{\mathrm{tr}}
\newcommand{\Lrows}{\bb{L}_{\mathrm{r}}}
\newcommand{\Lcols}{\bb{L}_{\mathrm{c}}}
\newcommand{\Lambdarows}{\bb{\Lambda}_{\mathrm{r}}}
\newcommand{\Lambdacols}{\bb{\Lambda}_{\mathrm{c}}}
\newcommand{\Grows}{\mathcal{G}_{\mathrm{r}}}
\newcommand{\Gcols}{\mathcal{G}_{\mathrm{c}}}

\newcommand{\Ga}{\mathcal{G}_1}
\newcommand{\Gb}{\mathcal{G}_2}
\newcommand{\G}{\mathcal{G}}

\newcommand{\Ez}{E_{\mathrm{data}}}

\newcommand{\Ereg}{{E}_{\mathrm{reg}}}

\newcommand{\Edirrows}{{E}^{r}_{\mathrm{dir}}}
\newcommand{\Edircols}{{E}^{c}_{\mathrm{dir}}}

\newcommand{\Ediagrows}{{E}^{r}_{\mathrm{diag}}}
\newcommand{\Ediagcols}{{E}^{c}_{\mathrm{diag}}}

\newcommand{\dirwtr}{\mu_r}
\newcommand{\dirwtc}{\mu_c}
\newcommand{\diagwtr}{\rho_r}
\newcommand{\diagwtc}{\rho_c}

\newcommand{\squarespace}{\square\,}
\renewcommand{\eqref}[1]{(\ref{#1})}

\usepackage{xcolor}
\usepackage{soul}

\title{Geometric Matrix Completion: A Functional View}

\author{%
 Abhishek Sharma\\
LIX, \'Ecole Polytechnique\\
{\tt\small kein.iitian@gmail.com}
\and
Maks Ovsjanikov\\
LIX, \'Ecole Polytechnique\\
{\tt\small maks@lix.polytechnique.fr}
}
\begin{document}

\maketitle

\begin{abstract}
  We propose a totally functional view of geometric matrix completion problem. Differently from existing work, we propose a novel regularization inspired from the functional map literature that is more interpretable and theoretically sound. On synthetic tasks with strong underlying geometric structure, our framework outperforms state of the art by a huge margin (two order of magnitude) demonstrating the potential of our approach. On real datasets, we achieve state-of-the-art results at a fraction of the computational effort of previous methods. Our code is publicly available at \url{https://github.com/Not-IITian/functional-matrix-completion}
\end{abstract}
\section{Introduction}Matrix completion deals with the recovery of missing values of a matrix of which we have only measured a subset of the entries,
\begin{equation}\label{eq:matrix_completion}
    \mathrm{Find}\;\;\Mhat\;\;\mathrm{s.t.}\;\; \Mhat\odot\mask = \M \odot \mask.
\end{equation}
Here $\Mhat$ stands for the unknown matrix, $\M\in\mathbb{R}^{m\times n}$ for the ground truth matrix, $\mask$ is a binary mask representing the input support, and $\odot$ denotes the Hadamard product. In general, without any constraints, this problem \eqref{eq:matrix_completion} is ill-posed and not solvable. However if the rank of underlying matrix is small, the number of degrees of freedom decreases and thus, it is common to find the lowest rank matrix that agrees with known measurements. Under this low rank assumption, the matrix completion problem can be rewritten as,
\begin{equation}\label{eq:minRankLS}
    \min_{\Mhat}\; \mathrm{rank}\left(\Mhat\right)+\frac{\mu}{2}\left\|\left(\Mhat- \M\right) \odot \mask\right\|_F^2.
\end{equation}

 Various problems in collaborative filtering can be posed as a {\em matrix completion} problem~\cite{kalofolias2014matrix,rao2015collaborative},  where for example the columns and rows represent users and items, respectively, and matrix values represent a score determining whether a user would like an item or not. This setting was particularly popularized by the Netflix challenge \cite{Koren09}. Often, additional structural information is available in the form of column and row graphs representing similarity of users and items, respectively. Such geometric information is not exploited by rank based prior work that only seeks a purely algebraic solution by optimizing for low rank~\cite{candes2009exact}. Prior work that incorporates geometric structure into matrix completion problems\cite{monti2017geometric} obtains state of the art results using powerful pattern extraction ability of graph CNN but falls short of giving a principled framework to model such geometric information.  

\cite{dmf_spectral20} makes an attempt to build a principled framework that is based on a functional map representation~\cite{ovsjanikov2012functional} and also compete empirically with highly engineered models such as multi-graph CNN\cite{monti2017geometric}. One of the advantages of working with the functional map representation  is that its size is typically much smaller, and is only controlled by the size of the basis, independent of the number of nodes in graphs, resulting in simpler optimization problems
Although \cite{dmf_spectral20} obtains state-of-the-art results on both synthetic and real datasets, it introduces several non-convex regularization terms thereby, making the overall optimization  harder to optimize. To address this challenge, we propose a simple formulation based on functional map, consisting of a single regularizer, that mitigates the problems associated with \cite{dmf_spectral20}.

\paragraph{Contributions.}
 Our contributions are threefold. First, we propose a 
 novel functional view of geometric matrix completion problem that is convex in formulation and theoretically grounded. Second, our method, on synthetic tasks with strong underlying geometric model, sets new benchmarks in modelling the geometric information that are $100$ times superior over prior work. Third, our proposed model obtains state-of-the-art results on various real world recommendation systems datasets while being more intuitive and  simpler to optimize and thereby,  easier to analyze and reproduce. 

\section{Related work}
Matrix completion has been studied with many viewpoints and thus, exhaustive coverage of prior work is beyond the scope of this paper. In this section, we mainly describe related work on geometric matrix completion.  

\paragraph{Geometric matrix completion.}
A prominent relaxation of the rank operator in Eq. \eqref{eq:minRankLS} is to constrain the space of solutions
to be smooth w.r.t. some geometric structure of the
matrix rows and columns. There exist several prior work on geometric matrix completion problem that exploit such geometric information \cite{berg2017graph,kalofolias2014matrix,rao2015collaborative} such as graphs encoding relation between rows and columns. More recent work leverages deep learning on geometric domain \cite{berg2017graph,monti2017geometric} to extract relevant information from geometric data such as graphs.  As argued in \cite{dmf_spectral20}, while these techniques achieve state-of-the-art results, their design is highly engineered and thus, non-intuitive. 

\paragraph{Functional Maps.}
Our work is mainly inspired from the functional map framework~\citep{ovsjanikov2012functional} used ubiquitously in non-rigid shape correspondence, and has been extended to handle challenging partial matching cases, e.g. ~\cite{litany2017fully}. This framework has recently been adapted for geometric matrix completion in~\cite{dmf_spectral20}, where the authors propose to build a functional map between graphs of rows and columns. As noted in several works, isometry between two spaces is a key to functional map representation. Assuming isometry between real world graphs is however over optimistic. Thus, one way to work under relaxed isometry condition is to instead align the eigen basis with additional transformation matrix to achieve diagonal functional map matrix\cite{litany2017fully}. \cite{dmf_spectral20} achieve this with a range of transformation on eigen basis of graph Laplacian. However, they 1) impose several regularization terms each with a scaling hyperparameter and some even with different initialization 2) explore a huge range of hyperparameter space and question remains on why does it work so well.
\section{Preliminaries}
\label{Preliminaries}
In this section, we cover some preliminaries about product graphs and functional maps.

\paragraph{Product graphs}Let $\G = (V,E,W)$  be a (weighted) graph  with its vertex set $V$ and edge set $E$ and adjacency matrix denoted by $W$. Graph Laplacian $\bb{L}$ is given by $\bb{L} = \bb{D}-\bb{W}$,
where $\bb{D} = \mathrm{diag}(\bb{W}\bb{1})$ is the \textit{degree matrix}. $\bb{L}$ is symmetric and positive semi-definite and therefore admits a spectral decomposition $\bb{L} = \Phibold\bb{\Lambda}\Phibold^\top$. It is well-known  that spectrum of the Laplacian contains the structural information about the graph \cite{spielman2009spectral}. Let $\Ga = \left(V_1,E_1,W_1\right)$, $\Gb = \left(V_2,E_2,W_2\right)$ be two graphs, with $\bb{L}_1= \Phibold\bb{\Lambda}_1\Phibold^\top$, $\bb{L}_2 = \Psibold\bb{\Lambda}_2\Psibold^\top$ being their corresponding graph Laplacians. The bases $\Phibold,\Psibold$ can be used to represent functions on these graphs. We define the Cartesian product of $\Ga$ and $\Gb$, denoted by $\Ga\squarespace \Gb$, as the graph with vertex set $V_1\times V_2$, on which two nodes $(u,u'),(v,v')$ are adjacent if either $u=v$ and $(u',v')\in E_2$ or $u'=v'$ and $(u,v)\in E_1$.

\paragraph{Functional maps.}Let $\Mhat$ be a function defined on $\Ga\squarespace \Gb$.  It can be encoded as a matrix of size or $|V_1| \times |V_2|$. Then it can be represented using the bases $\Phibold,\Psibold$ of the individual graph Laplacians, $\bb{C} = \Phibold^\top\Mhat\Psibold$. In the shape processing community, such $\bb{C}$ is called a \textit{functional map}, as it it used to map between the functional spaces of $\Ga$ and $\Gb$.  One of the advantages of working with the functional map representation $\bb{C}$ rather than the matrix $\bb{X}$ is that its size is typically much smaller, and is only controlled by the size of the basis, independent of the number of nodes in $G_1$ and $G_2$, resulting in simpler optimization problems. Moreover, the projection onto a basis also provides a strong regularization, which can itself be beneficial for both shape matching, and, as we show below, matrix completion. For example, given two functions, $\bb{x} = \Phibold\bb{\alpha}$ on $\Ga$ and $\bb{y} = \Psibold\bb{\beta}$ on $\Gb$, one can use $\bb{C}$ to map between their representations $\bb{\alpha}$ and $\bb{\beta}$, i.e., $\bb{\alpha} = \Phibold^\top\bb{x}=\bb{C}\Psibold^\top\bb{y} = \bb{C}\bb{\beta}$.
\label{sec:background}
\section{Functional Geometric Matrix Completion}
We assume that we are given a set of samples from some unknown matrix $\bb{M}\in\mathbb{R}^{m\times n}$, along with a binary indicator mask $\bb{S}$ that is $1$ for measured samples and $0$ for missing ones. In addition, we are given two graphs $\Grows,\Gcols$, encoding relations between the rows and the columns, respectively. We represent the Laplacians of these graphs and their spectral decompositions by $\Lrows= \Phibold\Lambdarows\Phibold^\top$, $\Lcols = \Psibold\Lambdacols\Psibold^\top$.  We minimize the  objective function of the following form:
\begin{equation}\label{eq:initialOpt}
\begin{split}
    &\min_{\Mhat}\; \Ez(\Mhat)+\mu \Ereg(\Mhat)
\end{split}
\end{equation}
with $\Ez$ denoting a data term of the form
\begin{equation}
\label{eq:data}
\Ez(\Mhat) = \left\|\left(\Mhat- \M\right) \odot \mask\right\|_F^2,
\end{equation}
As observed in \cite{dmf_spectral20}, we can decompose  $\Mhat =\Phibold\bb{C}\Psibold^\top$.  Remarkably, the data term itself, as we show in our experiments later, when expressed through the functional map i.e.$\Mhat =\Phibold\bb{C}\Psibold^\top$ already recovers low-rank matrices and outperforms the approach of \cite{dmf_spectral20} on synthetic geometric experiments. Before we explain the choice and motivation of our regularizer $\Ereg$, we explain next why the data term itself already works remarkably well on rank constrained geometric problems.

\subsection{Low Rank Geometric Matrix Completion}
 Our first observation is that by using a reduced basis to represent a function $\Mhat$ on the product space $\Ga\squarespace \Gb$ already provides a strong regularization, which can be sufficient to recover a low rank matrix approximation from a sparse signal.
 
 Specifically, suppose that we constrain $\bb{X}$ to be a matrix such that $\bb{X} =\Phibold\bb{C}\Psibold^\top$ for some matrix $\bb{C}$. Note that if $\Phibold$ and $\Psibold$ have $k$ columns each then $\bb{C}$ must be a $k \times k$ matrix. We would like to argue that solving Eq. \eqref{eq:data} under the constraint that $\bb{X} =\Phibold\bb{C}\Psibold^\top$ will recover the underlying ground truth signal  $\bb{Y}$ if it is low rank and satisfies an additional condition that we call basis consistency. 
 
For this suppose that the ground truth hidden signal $\bb{M}$ has rank $r$. Consider its singular value decomposition $\bb{M} = \bb{U} \bb{\Sigma} \bb{V}^\top $. If $\bb{M}$ has rank $r$, then $\bb{\Sigma}$ is a diagonal matrix with $r$ non-zero entries. We will call $\bb{M}$ \emph{basis-consistent} with respect to $\Phibold,\Psibold$ if the first $r$ left singular vectors $U_{r}$ (i.e., those corresponding to non-zero singular values) lie in the span of $\Phibold$, and the first $r$ right singular vectors $V_{r}$ lie in the span of $\Psibold$.
In other words, there exist some matrices $\bb{R},\bb{Q}$ s.t. $U_{r} = \Phibold \bb{R}$ (note that this implies $k \ge r$) and $V_{r} = \Psibold \bb{Q}$.

Given this definition, it is easy to see that all basis-consistent matrices with rank $r\le k$ can be represented by some functional map $\bb{C}$. In other words, given $\bb{Y}$ that is basis-consistent, there is some functional map $\bb{C}$ s.t. $Y = \Phibold \bb{C} \Psibold^T $. Conversely any $\bb{X} = \Phibold \bb{C} \Psibold^T$ has rank at most $k$ and must be basis-consistent by construction.

Second, suppose we are optimizing Eq \eqref{eq:data} under the constraint $\bb{X} =\Phibold\bb{C}\Psibold^\top$ and that the optimum, i.e., the ground truth matrix $\bb{M}$, is basis-consistent. Then since the energy $E(C)$ is convex and there are enough known samples to full constrain the corresponding linear system, then we are guaranteed to recover the optimum low-rank basis-consistent matrix. We note briefly that the argument above can also be made approximate, when the ground truth matrix is not exactly, but only approximately basis consistent, by putting appropriate error bounds.

This simple observation suggests that by restricting $\bb{X} = \Phibold\bb{C}\Psibold^\top$ and optimizing over the matrices $\bb{C}$ instead of $\bb{X}$ already provides a strong regularization that can help recover appropriate low-rank signals even without any other regularization. In practice, we observe that a weak additional regularization is often sufficient to obtain state-of-the-art results.

\subsection{Functional Regularization}
For clarity, we first describe the regularization terms introduced in \cite{dmf_spectral20} briefly 
\begin{equation}
\Ereg =  \dirwtr\Edirrows + \dirwtc \Edircols +\diagwtc\Ediagcols + \diagwtr\Ediagrows
\label{eq:sgmc_reg}
\end{equation}

$\Edirrows$ is the Dirichlet energy of $\Mhat$ on row graph, given by $\Edirrows(\Mhat) =\trace\left(\Mhat^\top\Lrows\Mhat\right)$. Similar term is used for column graph
$\Edircols(\Mhat)=\trace\left(\Mhat\Lcols\Mhat^\top\right)$    
In addition to Dirichlet energy, \cite{dmf_spectral20} also introduces two regularization on the transformation matrix $\bb{P}, \bb{Q}$. As described earlier, the purpose of these transformation is to rotate the original eigen basis $\Phibold\,\Psibold $ that will simplify the structure of $\bb{C}$. 
$\Ediagrows\left(\Rphi\right) = \|\mathrm{off}\left(\Rphi^\top\Lambdarows\Rphi\right)\|_F^2,$
where $\mathrm{off}(\cdot)$ denotes the off-diagonal elements. A similar treatment to the columns graph gives,$\Ediagcols\left(\Rpsi\right) = \|\mathrm{off}\left(\Rpsi^\top\Lambdacols\Rpsi\right)\|_F^2.$.  We briefly mention that in addition to SGMC,  \cite{dmf_spectral20} also proposes a multi-resolution spectral loss named SGMC-Zoomout(SGMC-Z)\cite{melzi2019zoomout} with its own hyperparameters (step size between different resolutions) besides the four hyperparameters in Eq.  \ref{eq:sgmc_reg}. 

In contrast to SGMC or SGMC-Z, our $\Ereg$ contains a single regularization term on the functional map induced between row space and column space described next.   

\paragraph{Laplacian Commutativity as a Regularizer}
Our main idea is to use the simplest possible regularizer, which furthermore leads to a convex optimization problem and can achieve state-of-the-art results. For this we borrow a  condition that is prominent in the functional map literature \cite{ovsjanikov2016computing}. Namely, in the context of surfaces, the functional map is often expected to \emph{commute with the Laplace-Beltrami operator}:
\begin{equation}
\Ereg = \big\Vert \bb{C}\Lambdarows - \Lambdacols \bb{C} \big\Vert^2,
\label{eq:ours_reg}
\end{equation}
 where $\Lambdarows$ and $\Lambdacols$ are diagonal matrices of Laplacian eigenvalues of the source graph (row graph) and target graph (column graph).

For shape matching problems, this constraint helps to find better mappings because functional maps that commute with the Laplacian must arise from near isometric point-to-point correspondences \cite{rosenberg1997laplacian,ovsjanikov2012functional}. More broadly, commutativity with the Laplacian imposes a diagonal structure of the functional map, which intuitively promotes preservation of low frequency eigenfunctions used in the basis. In the context of matrix completion this can be interpreted simply as approximate preservation of global low frequency signals defined on the two graphs.

Given these above definitions, our problem defined in Eq. \eqref{eq:initialOpt} reduces to 
\begin{equation}
\min_{\bb{C}} \left\|\left(\Mhat- \M\right) \odot \mask\right\|_F^2 + \mu \big\Vert \bb{C}\Lambdarows - \Lambdacols \bb{C} \big\Vert^2, \text{ where } \Mhat=\Phibold\bb{C}\Psibold^\top
\end{equation}

In practice, however, we observe faster convergence if we replace C with $\bb{P}\bb{C}\bb{Q}^\top$ and therefore, let all three be free variables.



\subsection{Implementation} The optimization is carried out using gradient descent in Tensorflow \cite{tensorflow2015-whitepaper}. 
\paragraph{Initialization} Similar to \cite{dmf_spectral20}, we initialize the $\bb{P}$ and  $\bb{Q}$ with an identity matrix with size equal to that of underlying matrix $\bb{M}$ corresponding to respective dataset and $\bb{C}$ by projecting $\Mhat\odot \mask$ on the first eigen vector of $\Lcols$ and $\Lrows$.

\paragraph{Hyperparameters} Our formulation contains two hyperparameters namely the size of $\bb{C}$ and the weighing scalar $\mu$. We  divide the number of available entries in the matrix randomly into training and validation set in a $95$ to $5$ ratio respectively. We set $\mu$ to be $.00001$ and learning rate to $.000001$ for all the experiments. Size of C is different for different datasets and set according to the performance on the  validation set of each dataset.

\section{Results}\label{sec:exp_study}
This section is divided into two subsections. 
 The goal of first subsection is to extensively compare between our approach and Spectral geometric matrix completion  \textbf{(SGMC)}\cite{dmf_spectral20} on a synthetic example of a community structured graph. In the second subsection, we compare with all approaches on various real world recommendation benchmarks. Note that we use SGMC and \cite{dmf_spectral20} interchangeably in this section.
    
\subsection{Experimental study on synthetic dataset}
For a fair comparison with \cite{dmf_spectral20}, we use the graphs taken from the synthetic Netflix dataset. Synthetic Netflix is a small synthetic dataset constructed by \cite{kalofolias2014matrix} and \cite{monti2017geometric}, in which the user and item graphs have strong communities structure.  It is useful in conducting controlled experiments to understand the behavior of geometry-exploiting algorithms. In all our tests, we use a randomly generated band-limited matrix on the product graph $\Gcols\squarespace\Grows$. 
\paragraph{Baselines}
\begin{itemize} 
\item{\textbf{Ours-FM}}; This baseline only optimizes for $\bb{C}$ without any regularization. All results are obtained with $\bb{C}$ of size $30 \times 30$.    
\item{\textbf{SGMC}}: All results are obtained with their open source code with their optimal parameters.
\end{itemize}
\begin{table}
\caption{Comparative results to test the dependence of SGMC and our method on the rank of the underlying random matrix of size $150 \times 200$}
\begin{center}
\begin{tabular}{|l|ccc|}
\hline
Rank  &  Ours  & Ours-FM& SGMC \\
\hline\hline

 5     & 1e-7 & 2e-5&1e-4 \\
    
10     &  2e-7 & 2e-5&2e-4\\
12         & 5e-7& 4e-5 &9e-4 \\
15      & 6e-3& 1e-3&1e-2 \\
20      & 3e-2& 1e-2 &5e-2\\
\hline
\end{tabular}
\end{center}
\label{table:res1}
\end{table}

\begin{table}
\caption{Comparative results to test the dependence of SGMC and our method on the  density of the sampling set in $\%$ of the number of matrix elements, for a random rank $10$ matrix of size $150 \times 200$.}
\begin{center}
\begin{tabular}{|l|ccc|}
\hline
Density &  Ours& Ours-FM & SGMC \\
\hline\hline

 1     & 2e-2 & 2e-2 &1e-1\\
    
5     &  8e-7 & 1e-3 &5e-4 \\
10         & 2e-7& 5e-5 &2e-4\\
20      & 1e-7 & 2e-5  &1e-4\\
\hline
\end{tabular}
\end{center}

\label{table:res2}
\end{table}

\begin{table}
\caption{Comparative results to test the robustness of our method in the presence of noisy graphs.}
\begin{center}
\begin{tabular}{|l|ccc|}
\hline
Noise &  Ours& Ours-FM &SGMC \\
\hline\hline
5     & 1e-3  & 2e-3 & 5e-3\\
10   &4e-3      & 3e-3  &1e-2\\
20      & 6e-3  & 6e-3 &1e-2\\
\hline
\end{tabular}
\end{center}
\label{table:res3}
\end{table}

\paragraph{Test Error.}
To evaluate the performance of the algorithms in this section, we report the \textit{root mean squared error},
\begin{equation}\label{eq:RMSE}
    \mathrm{RMSE}(\Mhat,\mask) = \sqrt{\frac{\left\|\left(\Mhat-\M\right)\odot\mask\right\|_F^2}{\sum_{i,j}\mask_{i,j}}}
\end{equation}
computed on the complement of the training set. Here $\bb{X}$ is the recovered matrix and $\mask$ is the binary mask representing the support of the set on which the RMSE is computed. 

We compare the two approaches on different constraints as follows:

\paragraph{Rank of the underlying matrix.}
We explore the effect of the rank of the underlying matrix, showing that as the rank increases upto $15$ to $20$, it becomes harder for both methods to recover the matrix.  As the rank increases, the reconstruction error increases, but it increases slower for us than for SGMC. For the training set we used $10\%$ of the points chosen at random (same training set for all experiments summarized in Table \ref{table:res1}). We remark that Ours-FM consistently outperforms SGMC for all rank.

\paragraph{Sampling density.}
We investigate the effect of the number of samples on the reconstruction error. We demonstrate that in the data-poor regime, our regularization is strong enough to recover matrix, compared to  performance achieved by incorporating geometric regularization through SGMC. These experiments are summarized in Table \ref{table:res2}. Note that gap between us and SGMC remains high even when the sample density increases to $20\%$. Even when using $1\%$ of the samples, we perform better than SGMC. We also remark that Ours-FM  outperforms SGMC only when density is sufficient.

\paragraph{Noisy graphs.}
We study the effect of noisy graphs on the performance. We follow the same experimental setup as  \cite{dmf_spectral20} and perturb the edges of the graphs by adding random Gaussian noise with zero mean and tunable standard deviation to the adjacency matrix. Table \ref{table:res3} mentions the value of this tunable standard deviation. We discard the edges that became negative as a result of the noise, and symmetrized the adjacency matrix. Table \ref{table:res3} demonstrates that our method is robust to  substantial amounts of noise in graphs. Surprisingly, Ours-FM demonstrates even stronger resilience to noise. 
\begin{table}
\caption{ Test error on Synthetic Netflix \citep{monti2017geometric}, Flixster \citep{jamali2010matrix}, and Movielens-100K \citep{harper2016movielens}}
\centering
\begin{tabular}{l r r r  r r r}
\textbf{Model} & $\substack{ \textbf{Synthetic}\\\textbf{Neflix}}$& \textbf{Flixster}   & \textbf{ML-100K}&
\\[0.05em] \\[-0.8em]
MC \citet{candes2009exact}& -- & $1.533$  & $0.973$ \\
GMC \citet{kalofolias2014matrix} & $0.3693$ & -- & $0.996$ \\
GRALS \citet{rao2015collaborative} & $0.0114$ & $1.245$  & $0.945$  \\
RGCNN \citet{monti2017geometric} & $0.0053$ & $0.926$  & $0.929$  \\
GC-MC \citet{berg2017graph} & -- & ${0.917}$  & ${0.910}$ \\
Ours-FM  & $0.0064$ &  $1.02$ &  $1.12$  \\
DMF \citet{arora2019implicit} & $0.0468$ & $1.06$ &   $0.922$  \\
SGMC & $0.0021 $ & $0.900$ & $0.912$  \\
SGMC-Z  & $0.0036 $ & $0.888$ & $0.913$ \\
Ours  & $0.0022$ & $0.888$ & $0.915$ \\
\end{tabular}
\label{table:results}
\vspace{-0.6cm}
\end{table}

\subsection{Results on recommender systems datasets}
In addition to synthetic Netflix, we also validate our method on two more recommender systems datasets for which row and column graphs are available. Movielens-100K \cite{harper2016movielens} contains ratings of $1682$ items by $943$ users whereas Flixter \citep{jamali2010matrix} contains ratings of $3000$ items by $3000$ users. All baseline numbers, except Ours-FM, are taken from \cite{monti2017geometric} and \cite{dmf_spectral20}.
\paragraph{Baselines}
\begin{itemize}
    \item \textbf{SGMC(Z)}: In addition to SGMC,  \cite{dmf_spectral20} also proposed a multi resolution spectral loss named SGMC-Zoomout. 
    \item \textbf{DMF}: This is a matrix factorization approach that was adapted for matrix completion tasks by  \cite{dmf_spectral20}. Note that this approach does not incorporate any geometric information.
    \item \textbf{Ours-FM}: This method only optimizes the data term, over $\bb{C}$, without any regularization. All results are obtained with  $\bb{C}$ of size $30 \times 30$.   
\end{itemize}
We explain several observations from Table \ref{table:results}: First,  our baseline, Ours-FM, obtains surprisingly good performance across datasets. This underscores the regularization brought in by the laplacian eigen basis of row and column graphs. Second, non geometric model such as DMF shows competitive performance with all the other methods on ML-100K. This suggests that the geometric information is not very useful for this dataset. Third, our proposed algorithm is competitive with the other methods while being simple and interpretable. On Synthetic Netflix, we  obtain best results with randomization on underlying graph structure. We explore its effect in detail in supplement.  Furthermore, it should be noted that non geometric models such as DMF performs poorly on both  synthetic datasets compared to ours and SGMC. Lastly, these experimental results validate the effectiveness of our single regularization when compared to the combination of several regularizations introduced in \cite{dmf_spectral20}.

\paragraph{Computation Issues} Our method depends 
on the eigenvalue decomposition of graph Laplacian matrix which is the main bottleneck to scale our approach for large scale deployment. We intend to address this issue in our future work.
 
\section{Conclusion}  In this work, we propose a functional view for geometric matrix completion, building upon the recent work of \cite{dmf_spectral20}. We establish empirically and theoretically that using a reduced basis to represent a function on the product space of two graphs already provides a strong regularization, which is sufficient to recover a low rank matrix approximation from a sparse signal. Moreover, we propose a novel regularization and show, through extensive experimentation
on real and synthetic datasets, that our single regularization is very competitive when compared to the combination of several different regularizations proposed before.   
\section{Acknowledgement} 
Parts of this work were supported by the ERC Starting Grant StG-2017-758800 (EXPROTEA), KAUST OSR Award No. CRG-2017-3426, and a gift from Nvidia.

\bibliographystyle{plainnat}
\bibliography{geom_completion}

\end{document}